%% file: iclr2025_conference.tex
\documentclass{article} % For LaTeX2e
\usepackage{iclr2025_conference,times}

\input{math_commands.tex}

\usepackage{microtype}
\usepackage{graphicx}
\usepackage{booktabs} % for professional tables

\usepackage{hyperref}
\usepackage{url}
\usepackage{multirow}
\usepackage{array}

\usepackage{amsmath}
\usepackage{amssymb}
\usepackage{mathtools}
\usepackage{amsthm}
\usepackage{graphicx}   % For adding images
\usepackage{subcaption} % For subfigure formatting (optional)
\usepackage{caption}    % For better caption formatting
\usepackage{enumitem}
\usepackage[capitalize,noabbrev]{cleveref}
\usepackage{lipsum}
\input{notations.tex}
\definecolor{outline}{RGB}{0,0,200}
\definecolor{comment}{RGB}{200,0,0}
\usepackage{titlesec}

\title{Optimizing GPT for Video Understanding: Zero-Shot Performance and Prompt Engineering}

\author{Mark Beliaev,\  Victor Yang,\  Madhura Raju,\  Jiachen Sun,\  Xinghai Hu \\
Tiktok Inc.\\
}

\iclrfinalcopy % Uncomment for camera-ready version, but NOT for submission.
\begin{document}
\maketitle
\input{sec/0_abstract}    
\input{sec/1_intro}
\input{sec/2_formatting}

\end{document}

%% file: math_commands.tex
\usepackage{amsmath,amsfonts,bm}

\def\eqref#1{equation~\ref{#1}}
\def\1{\bm{1}}

\DeclareMathAlphabet{\mathsfit}{\encodingdefault}{\sfdefault}{m}{sl}
\SetMathAlphabet{\mathsfit}{bold}{\encodingdefault}{\sfdefault}{bx}{n}

%% file: notations.tex
\newcommand{\policy}{\{POLICY\}}
\newcommand{\category}{\{CATEGORY\}}

%% file: sec/0_abstract.tex
\begin{abstract}
In this study, we tackle industry challenges in video content classification by exploring and optimizing GPT-based models for zero-shot classification across seven critical categories of video quality. We contribute a novel approach to improving GPT’s performance through prompt optimization and policy refinement, demonstrating that simplifying complex policies significantly reduces false negatives. Additionally, we introduce a new decomposition-aggregation-based prompt engineering technique, which outperforms traditional single-prompt methods. These experiments, conducted on real industry problems, show that thoughtful prompt design can substantially enhance GPT’s performance without additional finetuning, offering an effective and scalable solution for improving video classification.\end{abstract}

%% file: sec/1_intro.tex
\section{Introduction}
\label{sec:intro}
  In the past decade, recognition over mutliple modalities has become an increasingly critical challenge for large video platform companies, such as TikTok and YouTube, which rely on managing vast amounts of user-generated content. Unlike traditional classification tasks that process single-modality inputs, multi-modal classification combines multiple data sources, such as image, audio, and text, to provide more accurate and context-aware predictions~\cite{wang2017truly}. The ability to reason across these different modalities is essential for effective content classification. This capability is particularly important for identifying inappropriate content, or videos that do not align with platform policies, as it ensures a more nuanced understanding of content beyond just visual or textual cue~\cite{de2019multimodal, yousaf2022deep}.

With the advent of large language models (LLMs) like GPT-4, a new paradigm has emerged in the field of multi-modal classification~\cite{achiam2023gpt}. LLMs offer the promise of generalization, boasting the ability to perform a wide range of tasks with minimal task-specific data, often through one-shot or few-shot learning. Their large-scale pretraining on diverse datasets imbues them with a form of “world knowledge” that can be flexibly adapted to a range of problems, including those that involve multi-modal inputs. As a result, generative AI  models are now being explored as a promising solution for problems traditionally addressed by multi-modal architectures~\cite{tang2023video}. One domain ripe for this exploration is video classification, where the challenge lies in understanding video content through a combination of visual, auditory, and sometimes textual cues.

However, while LLMs like GPT-4 offer broad generalization, applying them to real-world, industry-specific problems has proven difficult. These models are typically pretrained on massive public datasets, meaning that company-specific data—especially multi-modal data related to proprietary video content—is rarely seen during pretraining. As such, LLMs may struggle to capture the nuances of these specific tasks without further tuning~\cite{prottasha2024parameter, zheng2024fine, wang2024peer}. Additionally, despite the hype surrounding generative models, there is still skepticism in the industry regarding their return on investment~\cite{goldmansachs2024aiinvestment}.

In this study, we aim to address these challenges by evaluating the performance of GPT-4 in real-world, industry-relevant video classification tasks that require multi-modal understanding. Specifically, we explore the potential of GPT-4 for classifying TikTok video content based on a variety of criteria. Our work differs from traditional multi-modal models in two key ways: first, by focusing on the zero-shot and few-shot capabilities of GPT-4, we investigate whether large, pretrained generative models can be effectively applied to multi-modal tasks without extensive fine-tuning. Second, we explore the impact of policy refinement and prompt engineering on GPT-4’s performance, addressing one of the core limitations of LLMs in industry—namely, their difficulty in handling complex, highly specific policies like those used to moderate video content.

Our contributions are threefold. First, we present a comprehensive evaluation of GPT-4’s ability to handle real-world, multi-modal video classification tasks, comparing its performance to existing, in-house classification models that leverage multi-modal features. We show that GPT-4 can perform on par with these models in several categories, but faces challenges in more complex classifications, such as clickbait videos. Second, we introduce novel techniques to enhance GPT-4’s performance in these tasks. We find that shortening complex policy prompts improves GPT-4’s ability to classify videos accurately by reducing false negatives, and that prompt engineering—specifically, dividing tasks like clickbait detection into subcategories—can lead to significant performance gains. Lastly, our experiments are grounded in actual industry datasets and problems, making the results highly applicable to video classification tasks at scale.

    \begin{figure}[!t]
    \centering
    \includegraphics[width=2.6in]{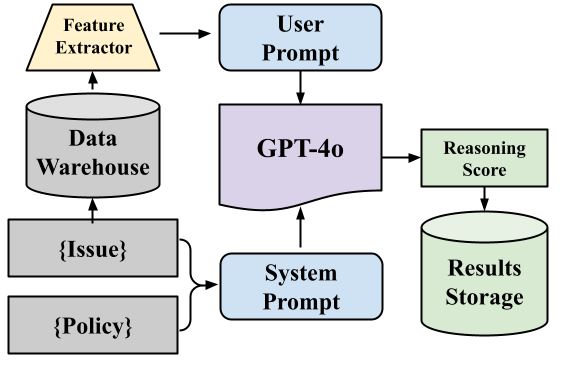}
    \caption{To have a fair comparison across categories, we design the experiment such that the item specific user prompt is independent of the category's policy, while the system prompt provided to GPT-4o incorporates the provided policy. For our experiments, given an \category~and \policy~along with a corresponding dataset, containing at least the \textit{item\_id} and ground truth \textit{label}, we ask GPT-4o to output a classification prediction, providing its \textit{reasoning} and \textit{score}.}
    \label{fig:exp_1}
\end{figure}

%% file: sec/2_formatting.tex
% \section{Formatting your paper}
\section{Related Work}\label{sec: Related Work}
    % {\color{outline} This should basically expand on Paragraph 3 of the introduction. It should be easy to use some of the related works I listed as a reference for more works. If you want, you can simply add:\par

    % "\noindent\textbf{Related Works}\quad Text goes here...",
    % into the introduction as Paragrah 3.}

    In this section, we review topics related to our study, including general and generative multi-modal models.
    
    \textbf{Multi-modal Classification}. Historically, multi-modal modeling has been approached using modality-specific architectures. For instance, BERT~\citep{devlin2018bert}, RoBERTa~\citep{liu2019roberta}, ResNet~\citep{he2016deep}, and ViT~\citep{dosovitskiy2020image} are commonly employed for text and vision tasks, respectively. Multi-modal models like CLIP~\citep{radford2021learning} and GLIP~\citep{li2022grounded} have effectively combined visual and textual inputs, leveraging correlations between modalities to achieve strong classification and detection results. However, these models typically require extensive task-specific fine-tuning and are limited by the rigid nature of their underlying architectures. This has resulted in a fragmented landscape where models are highly specialized for narrow tasks and lack generalizability to unseen problems or broader domains without significant retraining.
    
    \textbf{Large Generative Models}. In addition to classic tasks like recognition and detection, generative models have become a critical component of modern multi-modal learning, given that they are adaptable to different tasks. Recent advancements include models like Flamingo~\citep{alayrac2022flamingo}, OpenFlamingo~\citep{awadalla2023openflamingo}, LLaVA~\citep{liu2024visual}, InstructBLIP~\citep{dai2023instructblip}, and GPT-4o \citep{achiam2023gpt}, which integrate multiple modalities for content generation and reasoning tasks. Of particular interest for this study is GPT-4o, the state-of-the-art at the time of this project, which we utilize to evaluate content safety issues on video platforms. These generative models, unlike their classification counterparts, show greater adaptability and can be applied to a broader range of tasks with less task-specific tuning.

    % \lipsum[1-2]
    
    \section{Method}\label{sec: Problem Setup}
    This section details our methodology for testing the capability of GPT-4o at identifying various Feed Quality (FQ) categories across TikTok. We first provide some relevant background to the problem at hand in \cref{sec: Background}, following which we formulate how we test GPT-4o as a vision-language classifier in \cref{sec: Formulation}. We proceed by detailing our results and analysis in \cref{sec: Results}, and conclude by discussing some limitations and future directions in \cref{sec: Conclusion}.

    % \lipsum[1-2]
    \subsection{Problem Setup}\label{sec: Background}
    % {\color{outline} This subsection should describe the problem background in a way that is internally approved by the company. I have provided a draft which is in line with a typical academic publications, but does not regard any internal consideration from the company. I avoided using mathematical notation as we are not proposing any algorithm, and this seems typical for simple publications like this (its just binary classification).} 
    
We define the problem of identifying several domain specific categories across TikTok abstractly, as a binary classification task: given a sample TikTok video post, our goal is to identify whether the content is an example of the category at hand. We refer to posts which correspond to the category at hand as \textit{positive} cases, using the binary label $1$, while considering all other posts as \textit{negative} cases, using the binary label $0$. TikTok has For You feed eligibility standards that promote an appropriate experience for broad audiences by limiting sexually suggestive content, shocking \& graphic content, content that  tricks or manipulates others as a way to increase engagement, and unoriginal content. Each FQ category is defined by clear policy and guidelines that support consistent labeling for classification tasks. The resulting human annotations serve multiple purposes, including evaluating and monitoring algorithms aimed at identifying these categories.\par 

    % {\color{outline} On a second read through, I think the following paragraph addressing the challenges should be somehow described abstractly in the introduction (with some general background on FQ categories. Instead, we can use this section for defining the categories we consider + the baselines we use at TikTok. I have not done this, and not sure if it is better.} 
    %Such a system could address both challenges, alleviating the time constraint present when training annotators for a new category, and providing additional data that can be used for various monitoring and evaluation purposes. 
    
    Two major challenges to this problem are: (1) Training high quality annotators is difficult and time consuming. (2) Using solely human annotation for monitoring is impractical to support millions of daily posts. To this end, it is desirable to have intelligent systems which can perform well at the defined classification task across a plethora of FQ categories, preferably requiring little to no training. 
    Hence with the recent releases of LLMs equipped with the ability to process multi-modal inputs, it is natural to question whether we can utilize these off the shelf models for our task. Particularly, we are interested in the performance of GPT-4 at classifying TikTok videos according to the defined text policies, treating it as black-box classifier, with no additional training. Such a result would not only provide us with insight on GPT-4o's vision-language understanding, but also serve as a comparison for application specific models currently employed, as well as a baseline for more costly approaches requiring additional training.
    % \par  

    \subsection{Formulation}\label{sec: Formulation}
    In this section, we detail how we implement GPT-4 as a vision-language classifier for FQ categories in our experiments, and refer the reader to the official API used for reference~\cite{openai}. We provide a high level overview of our experiment design in~\cref{fig:exp_1}. Given a FQ category defined by the text strings \category, the name of the category, and \policy, the detailed text policy for the corresponding category, we provide GPT-4o with the following \textit{system} level instructions:
    
    \noindent\makebox[\linewidth]{\rule{\linewidth}{0.4pt}}
        \#\# Task: Label Videos for \category Content\\
        \#\#\# Objective:\\
        You are required to classify videos based on if they are \category content. The classification will help in training models to identify \category content. Each video should be labeled with an associated score indicating the likelihood of the video being \category content.\\ 
        \#\#\# Detailed Policy:\\
        \#\#\#\# \category:\\
        \policy\\
        \#\#\# Output:\\
        For each video, output a clear reasoning behind your decision and a score (0-100) indicating the overall likelihood of the video being \category content.\\
        Format your output as a JSON object with the following keys:\\
            reasoning: a chain of reasoning that explains how you arrived at your classification.\\
            score: An integer score from 0 to 100 representing how likely it is that the video is \category content, with 100 indicating that the video certainly is \category content, and 0 indicating that you are confident this is not \category.\\
        
        \#\#\# Notes:\\
        - Be clear and specific in your classification.\\
        - Inspect every frame provided. \\
        - Use the detailed policy to guide your judgment.\\
    \noindent\makebox[\linewidth]{\rule{\linewidth}{0.4pt}}

    Following this, for each TikTok video post we extract a set of text features: (audio transcription, hashtags, text, sticker text), alongside the video frames as base64 encoded images. Given this set of features, we query GPT-4o with the following \textit{user} level prompt, providing the video frames alongside:

    \noindent\makebox[\linewidth]{\rule{\linewidth}{0.4pt}}
    Given a video with the following features:\\
    \hspace*{.5cm}audio transcription: \{asr\}\\
    \hspace*{.5cm}hashtag: \{hashtag\}\\
    \hspace*{.5cm}text: \{text\}\\
    \hspace*{.5cm}sticker text: \{stickerText\}\\
    \hspace*{.5cm}video frames: images with base64 encoding provided\\
    Format your output as a JSON object with the specified keys.\\
    \noindent\makebox[\linewidth]{\rule{\linewidth}{0.4pt}}
    % \vspace{5}
    Based on our specified output format, we collect a JSON object containing the specified keys for each video, the reasoning used by GPT-4o as a text string, and the prediction score as an integer ranging from $0$ to $100$. 
    % {\color{outline} Can add point that Meng mentioned about lower stability of scores with floating points + asking GPT-4o\ to output reasoning \textbf{before} score.}
    	
    \section{Results}\label{sec: Results}
    Before providing a detailed analysis of the results, we give additional details on the setup of our experiments. 
    
    \subsection{Tasks}
     We conducted experiments using OpenAI’s latest GPT-4o model, accessed through Microsoft Azure at the time of the study: gpt-4o version 2024-05-13. While we also used Gemini-1.5-pro during preliminary experiments, we found the performance was similar and left it out when computing our final results. The temperature parameter was set to zero, with the top \(p = 1\) value fixed at one, without employing any stop-words. Additionally, the frequency penalty was set to 0.5, while the presence penalty remained at zero. %The Gemini version we compare in our preliminary study is Gemini-1.5-pro. 
     For each video, we sampled frames at $0.5$ frames per second, including the first and last frames, and used a maximum of 30 frames irrespective of the video length. 
    
    Using the formulation defined in the previous section, we test the performance of GPT-4o across $7$ FQ categories at TikTok: 
    \begin{itemize}
  \item \textbf{Sensitive and Mature Themes}:  Nudity and body exposure or sexually suggestive content
  \item \textbf{Shocking and Graphic Content}: Content that may intentionally shock, upset, or disgust others.
  \item \textbf{Non-Interactive Modules}: More than 2 content modules appear in one video, but there is no real communication, connection, or mutual response between them. 
  \item \textbf{Clickbait}: A tactic that tricks users to interact with TikTok video or accounts through follow, like, share, comment, finish and other actions in order to artificially get more traffic than they honestly would.
  \item \textbf{Static Frame (SF)}:  Video format but completely static, including pictures, solid color, screenshots.
  \item \textbf{Watermark}: The video includes watermarks of other social media platforms and apps.
  \item \textbf{Usefulness}: Content that conveys knowledge, experience, information that helps users to learn more.
  \end{itemize}

  % {\color{outline}We need a high level overview for each, should be in problem setup background section!} 
  For each category, we use a balanced dataset of at least $500$ video posts, and collect the reasoning and score provided by GPT-4o. Each video post has an associated ground truth label provided by a human annotator, and when available, a \textit{baseline} score, which corresponds to the normalized output of the current production model being used to identify this category at TikTok. Note that while the individual categories typically represent a small portion of the overall TikTok feed, in order to have a fair comparison across categories, we use balanced datasets containing equal number of positive and negative cases.

    \subsection{Baselines}
    For the aforementioned domain specific tasks, we used classification models trained on multi-modal features and human annotations as the baselines. The major difference between these baseline models and GPT-4o is that the baseline models are either rule-based or trained on the annotated samples, while GPT-4o is a much bigger model that is pretrained on large vision-language corpora. Additionally, for SF/Non-Interactive/Watermark, we could not collect enough training data to train a classifier by the time of this experiment. We tried rule-based method on these issues and only got acceptable performance on SF. Therefore, we omit reporting the baseline performance for Non-Interactive/Watermark in our experiment.
    % They are used to serve for all existing Tiktok videos and daily new videos, thus the model size is much smaller than GPT-4o. 

    \subsection{Experimental Setup}

\begin{figure}[!t]
		\centering
		\includegraphics[width=2.5in]{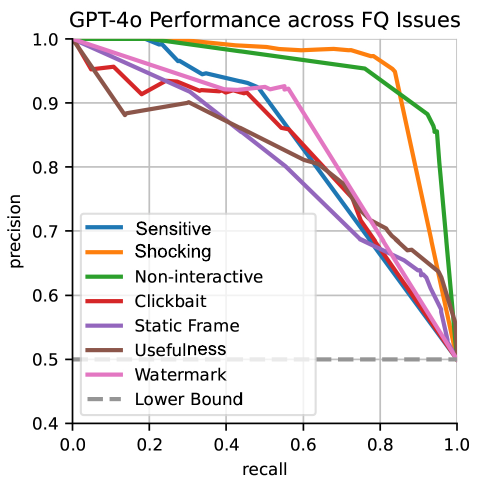}
		\caption{This figure shows the precision-recall curves when using the score provided by GPT-4o for all categories. Since all datasets are balanced to contain equal numbers of pos \& neg cases, they share one lower bound (0.5).}
		\label{fig:all_categories}
	\end{figure}

    Given these datasets, we perform three experiments which we overview below:
    
    \textbf{Exp. 1}\noindent\quad We test the performance of GPT-4o across the $7$ FQ categories defined, displaying the results in \cref{fig:all_categories} as precision-recall curves for each category, and providing additional statistics as well as a comparison to the baseline production models in \cref{tab:all_categories}. For all categories, we compare the AUC (Area Under the Curve) score, along with the total portion of samples that are false positives and false negatives, for both GPT-4o and the respective baseline model. We sort the categories by their AUC, provide the word count of each policy~(\#~words). For Non-Interactive/Watermark, the baselines are weak and thus we don't present the numbers here.

    \textbf{Exp. 2}\noindent\quad We perform an experiment on the Sensitive \& Mature category to test the affect of shortening the policy provided in the system prompt, displaying our results in \cref{fig:score_plot}.Specifically, we summarize the policy of Sensitive \& Mature category from 4023 words to XXX words, while comparing it with the Non-Interactive category at the same time, where the policy word number are just 116. 
    
    \textbf{Exp. 3}\noindent\quad We perform an experiment on the Clickbait category to test if we can further improve GPT-4o one-shot performance with simple prompting engineering, displaying our results in \cref{fig:exp_2_clickbait}. Specifically, we split the clickbait into 8 subcategories, ask GPT-4o to make independent prediction on these subcategories and then aggregate the scores into one combined prediction for this category.

	\begin{table}[!tb]
	\small
	\caption{GPT-4o vs. Baseline Models}
	% \raggedright 
	\label{tab:all_categories}
	\centering
	\setlength{\tabcolsep}{3pt}
	\begin{tabular}{cc|ccc|ccc}
		\toprule
		\multicolumn{2}{c}{Task} & \multicolumn{3}{c}{GPT-4o} & \multicolumn{3}{c}{Baseline}\\
		\cmidrule(r){1-2}\cmidrule(r){3-5}\cmidrule(r){6-8}
		% \cmidrule(r){1-1}\cmidrule(r){2-4}\cmidrule(r){5-7}
		category & \# words & AUC & FP & FN & AUC & FP & FN \\
            \midrule
            Non-Interactive & $116$ & $0.94$ & $0.08$ & $0.03$ & - & - & -\\
            Shocking & $934$ & $0.91$ & $0.01$ & $0.17$	& $0.97$ & $0.01$ & $0.12$ \\
            Usefulness & $143$ & $0.83$ & $0.18$ & $0.08$ & $0.90$ & $0.03$ & $0.22$\\
            Clickbait & $713$ & $0.79$ & $0.02$ & $0.30$ & $0.89$ & $0.02$ & $0.26$\\
            Static Frame & $80$ & $0.79$ & $0.26$ & $0.04$ & $0.79$ & $0.11$ & $0.13$\\ 
            Watermark & $402$ & $0.76$ & $0.02$ & $0.23$ &- & - & -\\
		Sensitive & $4023$ & $0.73$ & $0.00$ & $0.41$ & $0.95$ & $0.01$ & $0.25$\\
            \bottomrule
	\end{tabular}
	\end{table}
 
    % CAN BE SUBSECTION IF YOU WANT
    \subsection{Analysis}
    We analyze our research by answering the following $6$ questions sequentially, utilizing the aforementioned experiments.\par

    \begin{figure*}[!t]
		\centering
		\includegraphics[width=4.5in]{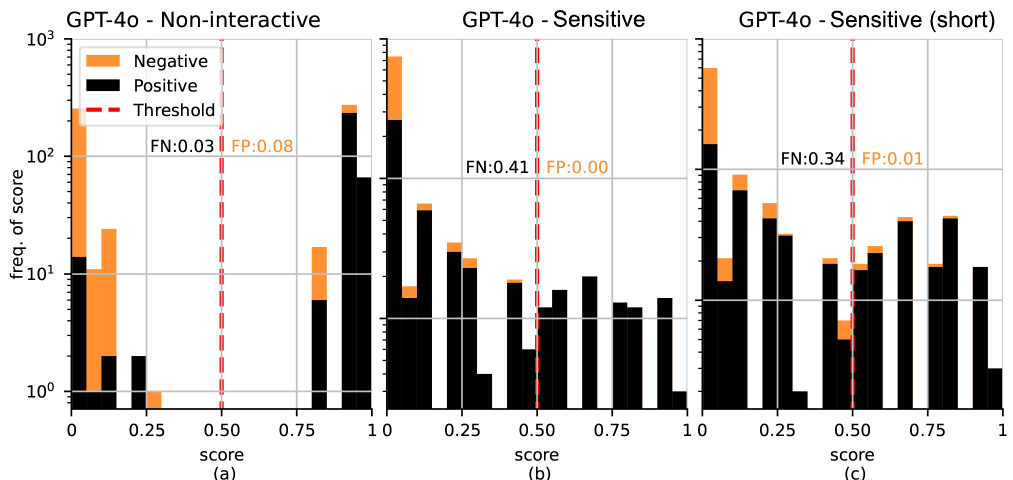}
		\caption{These charts plot the normalized (0-1) distribution of scores (x-axis) provided by GPT-4o, where the dashed red line is the classification threshold we choose, orange and black are used to depict negative and positive cases, and a logarithmic scale is used on the y-axis. We show the results for Non-interactive duet in (a), Sensitive and Mature Content in (b), and a shortened version of Sensitive and Mature Content in (c), which uses $96$ words in the policy instead of $4023$. We also report the total portion of False Negatives (FN) in the dataset: which are true positives (black), but classified as negative (left of the threshold), and conversely, the total portion of False Positives (FP), which are true negatives (orange), but classified as positive (right of the threshold).}
		\label{fig:score_plot}
	\end{figure*}

     \begin{figure}[!t]
		\centering
		\includegraphics[width=2.5in]{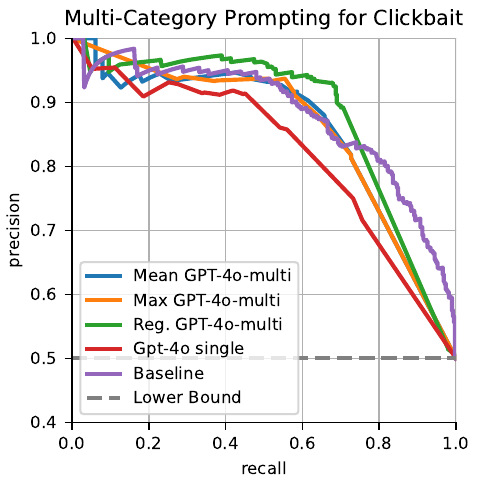}
		\caption{This figure shows the precision-recall curves for the Clickbait category when using both the original prompting technique (GPT-4o-single), as well as the proposed prompting technique which asks to provide a score for each category (GPT-4o-multi). To produce a final score for GPT-4o-multi, we consider the mean, max, as well as the best linear regression of the category scores. We additionally plot the corresponding production model (Baseline).}
		\label{fig:exp_2_clickbait}
	\end{figure}
 
     % Refer to main figure with all categories, talk about difference between several categories. 
    \textbf{Q1. How does GPT-4o perform at identifying FQ~categories?}\quad Using the results from \textbf{Exp. 1}, we take the score output by GPT-4o for each category, and plot the corresponding precision-recall curves in \cref{fig:all_categories} by sweeping through all possible classification threshold values. We see the best performing categories are Non-interactive and Shocking \& Graphic Content, while all other categories seem to be rather close in performance. This demonstrates a gap between certain categories which are simple to classify for GPT-4o, and categories which are more difficult. We explore this gap further in \textbf{Exp. 2} which is analyzed in \textbf{Q3-5} below. 

    \textbf{Q2. How does the one-shot performance of GPT-4o compare with current production models?}\quad We can see from our results of \textbf{Exp. 1} displayed in \cref{tab:all_categories} that GPT-4o performs on par with the production model baselines for Static Frame, Non-interactive, and Watermark. For Clickbait, Shocking \& Graphic Content, and Useful, the performance of GPT-4o is slightly lower ($\leq0.1$ AUC). For SD/SF/Watermark, since the baselines are rule-based due to lack of training data, GPT-4o shows the advantage on leveraging world knowledge and reasoning from generative pretraining. Finally, we note that GPT-4o shows the worst relative and absolute performance on Sensitive \& Mature. We continue our analysis by discussing what factors can contribute to the gaps in GPT-4o's performance between categories.\par 
    
    \textbf{Q3. What are the characteristics of categories that are difficult for GPT-4o?}\quad Looking closer at our results from \textbf{Exp. 1} by analyzing \cref{tab:all_categories}, we can make the following observations: 1) All baselines suffer performance due to False Negatives (FN) instead of False Positives (FP). 2) GPT-4o's predictions for Non-interactive, Useful, and Static suffer more from FP errors compared to FN errors. 3) GPT-4o's predictions for Clickbait, Watermark, Sensitive, and Shocking suffer more from FN errors compared to FP errors. The first observation is expected as production models are typically designed to be conservative, minimizing the affect that incorrect predictions have on the overall user experience. For the second and third point, we highlight that this pattern corresponds directly to the number of words in the policy: categories defined with less complex policies (which are easier to express in natural language), are dominated by FP errors instead of FN errors.\par 
    
    To further explore this last point, we can look at our results for \textbf{Exp. 2} displayed in \cref{fig:score_plot}. Looking first at the left plot (a), we see that the score output by GPT-4o for the Non-interactive category follows the desired trend: there is a valley in the middle, positive cases have high scores, and negative cases have low scores. The reported FP and FN are $0.08$ and $0.03$ respectively. Looking at the middle plot (b), we see the performance on Sensitive \& Mature is much worse: although FP is $0.00$, FN is $0.40$. The plot clearly shows that this is not just a trade off induced by the threshold choice, as there are many more positive (black) videos to the left of the threshold. Finally, looking at the right plot (c), we see the performance of $GPT-4o$ when we shorten the Sensitive \& Mature policy from $4023$ words to only $96$: the reported FN decreased by $0.07$ points, while FP increased by only $0.01$. The new AUC (not shown) using the short prompt is $0.79$, which is significantly larger than the previous $0.73$.\par 

\textbf{Q4. Why does the performance of $GPT-4o$ on Sensitive \& Mature improve when we shorten the policy?} We believe our results support the following hypotheses: when the policy is long and specific, GPT-4o is conservative - outputting a low score when it can not find correlations between the policy and the features. Since the policy is very detailed, GPT-4o paints with a fine brush, and is unlikely to mislabel negative videos as they do not correlate with the specific instructions in the policy (contributing to low FP). However, since many edge cases need to be considered for categories that require detailed policies, it is hard to find all positive cases (contributing to high FN). Conversely, when the policy is short, GPT-4o is aggressive - since a shorter policy has to make generalizations, GPT-4o paints with a broader brush and can identify more positive cases (contributing to low FN). However, negative videos are likely to be mislabeled due to this (contributing to high FP).\par

    \textbf{Q5. How can we extrapolate this result beyond Sensitive \& Mature Issues?}
    We conclude our discussion of \textbf{Exp 2.} by stating that these results do not provide a full story. We have to recall that shortening the prompt did not make most of the errors generated by FP cases, it only lowered the portion of FN cases. This is expected, we can identify more positive videos because we are lowering the specificity of what we consider as Sensitive \& Mature content. However, removing information from the prompt did not suddenly make the this category easier, like the Non-interactive Duet category. Although one may observe from our results that categories with large prompt size are dominated by FN errors, we can not say that this is due to prompt length alone. However, categories that inherently require a specific policy because they are hard to express in natural language due to the many edge cases present, can be characteristic of such behavior. Furthermore, long and detailed prompts are not solely responsible for poor performance, as observed by looking at the results for Shocking \& Graphic in \cref{tab:all_categories}.\par
    In conclusion, if we want good performance at identifying positive cases, we should consider optimizing the prompt length provided to GPT-4o, trying to express the category with as few words as possible, while still covering all relevant cases. Before concluding our work, we discuss some more ways to improve GPT-4o's one-shot performance.\par
    
    \textbf{Q6. What steps can we take to improve GPT-4o's one-shot performance?}\quad

    Throughout our study, we have tried several common methods for improving the one-shot performance of GPT-4o, such as manual prompt refinement and few-shot in-context instruction learning. Neither of these techniques provided significant performance improvement, and hence are not discussed here. However, we found that for the Clickbait category, which has a policy of $713$ words splitting the category into $8$ subcategories, asking GPT-4o to provide a score for each category gave favorable results. 
    % More details can be found in Appendix \ref{Sec: Additional Results}. 
    Specifically, we consecutively prompt GPT-4o to provide a reasoning and score for the corresponding category of Clickbait, and combine these $8$ scores by either taking the mean, the max, and the linear regression. We can see the results of \textbf{Exp. 3} in \cref{fig:exp_2_clickbait}, showcasing that all three techniques (GPT-4o-multi) produced better results than the previous method (GPT-4o-single). Interestingly, we found that performing linear regression on the scores allowed us to outperform the production model for higher precision values (which is the typical, more conservative, regime of interest.) 
    % {\color{outline} We note that these results should be taken lightly, as the regression was computed and evaluated on the same dataset. Nonetheless, when evaluating the individual correlations between category scores and the ground truth label, we saw that the largest correlation of $0.54$, was more than double the next largest correlation of $0.2$. (I just stated this result, but did not include the table. The data is below, showing the mean of the sore, correlation, coef of linear reg, and log loss. These are computed in my notebook.)} 
    Overall, we see that application specific prompt-engineering can lead to substantial performance improvement when using GPT-4o, without any additional training.\par 

% category	|mean	|corr (r2)	|coef	|logloss
% ------------------------------------------------------
% llm_score_1	|0.053	|0.164	|0.268	|15.97	|
% llm_score_2	|0.037	|0.196	|0.162	|15.13	|
% llm_score_3	|0.030	|0.179	|-0.005	|16.30	|
% llm_score_4	|0.046	|0.166	|-0.106	|14.02	|
% llm_score_5	|0.041	|0.200	|0.204	|15.96	|
% llm_score_6	|0.015	|0.103	|0.236	|17.01	|
% llm_score_7	|0.215	|0.546	|0.693	|7.14	|
% llm_score_8	|0.012	|0.112	|0.115	|16.27	|

    \section{Conclusion}\label{sec: Conclusion}

    In this paper, we investigate the application of GPT-4 on video content moderation, providing a strong baseline for the use of LLMs in multi-modal classification. By demonstrating the potential of generative models in solving this industry problem, we offer insights that can guide the development of more effective, scalable solutions for industries that rely on classifying multi-modal data. Our findings show that with the right adaptations, these models can be powerful and practical tools for industry-specific classification problems.
    %In this paper, we investigate the application of GPT-4o on video classification and provide a strong baseline on the application of LLMs in video classification and other multi-modal tasks. By demonstrating the potential of generative models in solving practical industry problems, we offer insights that can guide the development of more effective, scalable solutions for industries that rely on multi-modal data. Moreover, our findings challenge the notion that LLMs are not suited for industry-specific applications, showing that with the right adaptations, these models can be both powerful and practical tools for real-world classification problems.

    \textbf{Limitations:} 
    % We do not consider finetuning, as our focus is one shot performance. Additionally, 
    % we do not list results of {\color{red} similar} VLM models (e.g. Gemini-1.5-pro), {\color{red} since we} found that initial experiments showed very similar performance between them. {\color{red}Lastly, 
    We could not test against new \textit{reasoning}-models such as Deepseek's R1 or OpenAI's o1 and o3 models as they were not released at the time of this study. However, we believe that their increased capabilities do not outweigh the larger cost when considering the scale of content moderation in industry. Furthermore, prompt-engineering techniques, like the one introduced in Exp. 3, can be seen as a cost-effective way to inject \textit{reasoning} that is calibrated for the task at hand.
    
    \textbf{Future Directions:}  A clear direction for future works is finetuning. While performance increase is one benefit, inference cost is also important for industry applications. Therefore, clever design that integrates vision and text modalities with smaller, fine-tuned LLM's, is a promising direction. Furthermore, audio should be considered as an additional modality once pre-trained foundation models are capable of processing it alongside video and text. %Most future works will consider %: finetuning of GPT vision, finetuning Bert or VLM directly on these labels, and finetuning with text based, two stage finetuning approach. Another direction is to we can consider additional modalities such as sound, but also need to wait for advancement of GPT. 

    \bibliography{refs}
    \bibliographystyle{icml2024}